\documentclass{article}

\usepackage{arxiv}

\usepackage[utf8]{inputenc} % allow utf-8 input
\usepackage[T1]{fontenc}    % use 8-bit T1 fonts
\usepackage{hyperref}       % hyperlinks
\usepackage{url}            % simple URL typesetting
\usepackage{booktabs}       % professional-quality tables
\usepackage{amsfonts}       % blackboard math symbols
\usepackage{nicefrac}       % compact symbols for 1/2, etc.
\usepackage{microtype}      % microtypography
\usepackage{graphicx}
\usepackage{natbib}
\usepackage{doi}

\usepackage{amsmath, caption, subcaption, color, colortbl}
\definecolor{SpiritBlue}{rgb}{0.49,0.72,0.72}
\usepackage[onehalfspacing]{setspace}

\newcolumntype{C}[1]{>{\centering\arraybackslash}m{#1}}
\newcolumntype{L}[1]{>{\raggedright\arraybackslash}m{#1}}
\newcolumntype{N}{@{}m{0pt}@{}}
\title{Enhancing team performance with transfer learning during real-world human-robot collaborative learning}

\author{Athanasios C. Tsitos\\
    National Centre for Scientific Research \\
     `Demokritos' \\ 
     and \\
	University of Piraeus\\
	Greece \\
	\texttt{thtsitos@gmail.com} \\
	\And
	M. Dagioglou \\
	National Centre for Scientific Research \\
     `Demokritos' \\
     Greece \\
    \texttt{mdagiogl@iit.demokritos.gr}
    }

\date{}

\renewcommand{\headeright}{Tsitos et al.}
\renewcommand{\shorttitle}{Transfer learning for real-world HRC}

\hypersetup{
pdftitle={Transfer learning for real-world HRC},
pdfsubject={cs.RO, cs.AI, cs.HC},
pdfauthor={Athanasios C. Tsitos, M. Dagioglou},
pdfkeywords={human-robot co-learning, deep reinforcement learning, soft actor-critic, transfer learning, probabilistic policy reuse},
}

\begin{document}
\maketitle

\begin{abstract}
	Socially aware robots should be able, among others, to support fluent human-robot collaboration in tasks that require interdependent actions in order to be solved. Towards enhancing mutual performance, collaborative robots should be equipped with adaptation and learning capabilities. However, co-learning can be a time consuming procedure. For this reason, transferring knowledge from an expert could potentially boost the overall team performance. In the present study, transfer learning was integrated in a deep Reinforcement Learning (dRL) agent. In a real-time and real-world set-up, two groups of participants had to collaborate with a cobot under two different conditions of dRL agents; one that was transferring knowledge and one that did not. A probabilistic policy reuse method was used for the transfer learning (TL). The results showed that there was a significant difference between the performance of the two groups; TL halved the time needed for the training of new participants to the task. Moreover, TL also affected the subjective performance of the teams and enhanced the perceived fluency. Finally, in many cases the objective performance metrics did not correlate with the subjective ones providing interesting insights about the design of transparent and explainable cobot behaviour.
\end{abstract}

% keywords can be removed
\keywords{human-robot co-learning \and deep reinforcement learning \and soft actor-critic \and transfer learning \and probabilistic policy reuse}

%%%%%%%%INTRODUCTION%%%%%%%%%%%%
\section{Introduction}

As robots become part of our everyday life they are expected to resume several roles including that of the collaborator. In games \citep{sfikas2020collaborative,daronnat2020impact}, industrial set-ups \citep{kragic2018interactive, villani2018survey} and rehabilitation \citep{chiriatti2020framework}, humans and Artificial Intelligence (AI) agents, embodied or not, will collaborate to achieve common goals. Collaborative robots (cobots) particularly will interact in close proximity with humans, share goals and perform interdependent tasks \citep{Btepage2017HumanRobotCF}. Thus, they should be equipped with several `socially-aware' capabilities from human perception to co-learning and adaptation \citep{van2021becoming}. Taken together, these capabilities will have to support fluent, uninterrupted and natural human and team behaviour in a way that triggers social attitudes, similarly to  human-human interactions (HHI) \citep{sebanz2006joint}. 

Nevertheless, human-robot co-learning, similar to any de novo learning \citep{krakauer2019motor}, can be a time consuming procedure that depends on the motor and cognitive load demanded, the skills of the human partner, as well as the machine learning methods used and the computational complexity of the task. Naturally, cobots should be able to support reasonably fast training periods, as well as to integrate in their actions the capabilities of their human partners and adapt to their strengths and weaknesses. 

Recent advances in deep reinforcement learning (dRL) now allow us to study several aspects of human-robot collaboration (HRC) in real-time. Overall, dRL has been applied to many different robotic applications such as mobile platforms \citep{surmann2020deep, liu2020robot}, robotic arm control \citep{johannink2019residual, james20163d}, robotic grasping \citep{mohammed2020review, joshi2020robotic}, humanoid robots \citep{ozaln2019implementation, garcia2020teaching}, drones \citep{hodge2021deep, munoz2019deep}, quadruples \citep{haarnoja2018learning} and others. The success of the dRL in different robot environments lies in the ability of the dRL policies to learn motions and behaviours that, due to their complexity, are extremely difficult to be generated by hard-coded control laws.

A limitation of dRL in robotics is the generalization of knowledge in order to operate in new, unknown circumstances and environments, or with new collaborators \citep{nguyen2019review}. Most works try to solve a RL task by training the robot from scratch. However, this approach is non-optimal mainly because it is time consuming. In the case of the HRC, this is translated is long training periods that undermine the productivity of the team both in economical and fatigue-wise terms.

One way to overcome this limitation is by transferring knowledge. There are several ways that knowledge can be transferred in dRL frameworks \citep{zhu2020transfer}. In \textit{reward shaping}, the acquired knowledge is utilized to alter the reward function of the target task in order to accelerate training \citep{botteghi2020reward}. An issue here is related to changing  the convergence of the policy by altering the reward function. \cite{harutyunyan2015expressing} show how any function can be used to reshape the reward function while maintaining policy invariance.

Another technique for transfer learning (TL) is \textit{learning from demonstrations}. Here, the provided demonstrations encourage the agent to explore states which will help the convergence to an efficient policy faster. This can be achieved either with  `offline' \citep{zhang2018pretraining, schaal1996learning} or `online' \citep{hester2018deep} approaches. Naturally,  the quality of the demonstrated actions greatly affects the results.

In \textit{policy transfer} a previously learned policy is used to learn the new policy. One way to achieve this is by policy distillation \citep{rusu2015policy}, which means that the agent will select an action by minimizing the divergence of action distributions between the `teacher'(source) policies and the `student'(target) policy. Another approach is probabilistic policy reuse \citep{fernandez2006probabilistic}, where the agent can select an action based on the pre-learned policy instead of his own policy. For example, \cite{garcia2020teaching} use this to teach a humanoid robot how to walk fast by exploiting a policy that allows the robot to walk in a normal speed.

Deep RL has also been exploited in the context of HRC offering great opportunities for studying the behaviour of both human and AI agents in real-time and in real-world. Specifically, a dRL Soft-Actor Critic (SAC) agent has been recently used for human-robot real-time and real-world collaborative learning \citep{shafti2020real}. The authors presented a HRC task where a Universal Robots UR10 cobot and a human collaborate to guide a ball on a tray attached to the cobot from a starting to a goal position. Seven participants were trained with a lightly pre-trained agent for 80 trials of 40 seconds each. The results showed that the performance of the teams greatly depended on the human participant. When tested at the end of the training, some teams achieved considerably high scores while others seemed to have failed to learn how to collaboratively solve the task. However, when these participants interacted with an expert agent and agents that were exposed to well-performing participants, the teams achieved an improved performance. These results show that: a) as would be expected, different humans require different training intervals and b) team performance can be improved when expert knowledge is exploited. 

Based on the aforementioned observations, TL would be a natural choice for enhancing human-robot co-learning. To evaluate the enhancement in the performance, the human-AI \emph{team performance} and \emph{co-learning} instead of the isolated evaluation of each partner \citep{van2019six,chattopadhyay2017evaluating,carroll2019utility} should be taken into account. Moreover, both objective and subjective measures need to be considered during the evaluation of an HRC team  \citep{dragan2015effects}. 

In addition to measures such as fluency, contribution, safety and trust \citep{hoffman2019evaluating}, factors such as the attributed identity,  as well as the self (human) sense-of-agency and sense-of-control can also play a critical role during the HRC. In HHI, these attributes are affected by various factors including, the perceptual distinctiveness of each actor's actions, the competitiveness or complementarity of partners' roles an the fairness of reward distribution \citep{dewey2014phenomenology,le2020agents}. 

However, the human behaviour during HRC might differ compared to that during HHI. Current literature suggests that there are both similarities and differences \citep{kramer2012human}. For example, HRI appears to affect the sense-of-agency in a similar way compared to HHI \citep{ciardo2020attribution}. On the other hand, if the collaborative action includes kinesthetic cues the sense-of-agency is affected differently when temaing-up with a robot \citep{grynszpan2019sense}. Another factor that impacts the collaboration of humans with automated artificial systems is the ability for prediciton \citep{sahai2017predictive}. The results above stress the importance of further studying human perceptions during HRC in order to inform AI methods for achieving explainability and transparency \citep{vouros2022explainable}.

The goal of the present study was to endow a UR3 cobot in real-world and in real-time with TL capabilities and to study the impact of TL on the performance of the team, the speed of learning, as well as the impact of HRC on human’s subjective perceptions regarding the robot, the collaboration and their selves. 
The main contributions are: a) the design of a HRC task that is challenging enough yet simple to set-up and that provides a reasonable observation window, b) the integration of the dRL and the TL methods, c) the deployment of the real-world and real-time set-up, which includes the integration of the methods and the study process in the robot operating system (ROS), d) the observation of the effects of TL considering both objective and subjective measures through the study of seventeen human-robot teams.

%%%%%%%%Materials and methods%%%%%%%%%%%%

\section{Methods and Materials}

\subsection{Human-Robot Collaborative Task}\label{MM:HRCtask}

\begin{figure}[h]
   \begin{center}
    \includegraphics[scale=0.60]{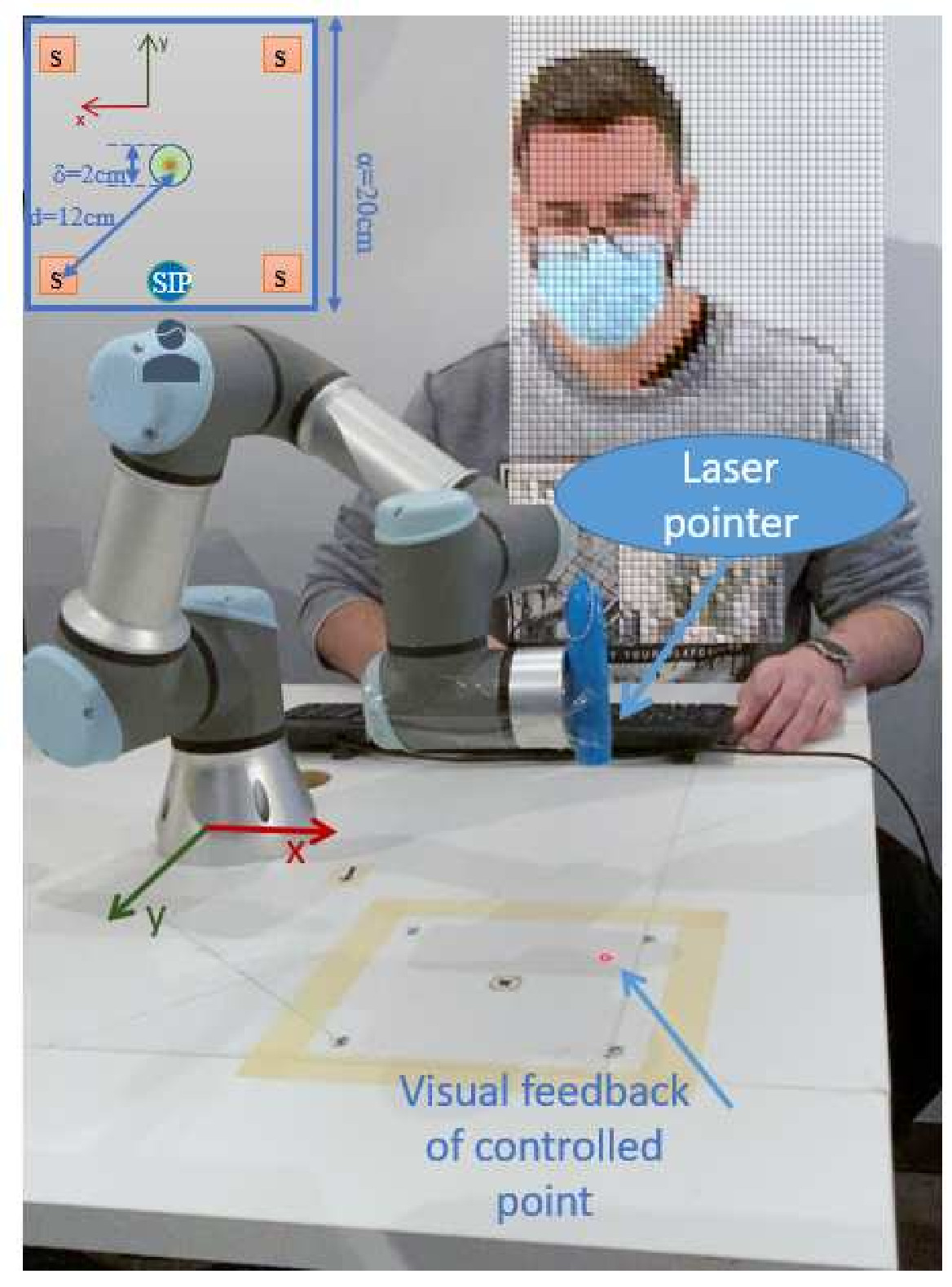}
    \end{center}
    \caption{The Human-Robot Collaboration setup. The robot's movements are constrained within a $20cm\times20cm$ area -  a schematic representation of the area is presented in the upper left corner of the figure. The EE is placed in one of the four starting (`S') positions and the HR team has to bring the EE in the centre (green area) of the square. A laser pointer attached to the EE of the robot provides to the human visual feedback about the position of the EE that is controlled.}
    \label{fig:setup}
\end{figure}

During the HRC task, a human and a dRL agent collaborate and control the end-effector (EE) of a Universal Robots UR3 cobot. The cobot's base is placed in the middle of a $1m\times1m$ table and its movement is constrained parallel to the table at a certain height (figure~\ref{fig:setup}). A laser pointer is attached to the EE in order to provide visual feedback about the EE's position to the human (red laser dot in figure~\ref{fig:setup}). The goal of the team is to jointly move the EE (the red laser dot) from a starting to a goal state (a goal position with a certain velocity). The starting point can be in one of the four corners of a $20cm\times20cm$ square (figure~\ref{fig:setup}), while the goal is always in the middle of the square. The human is responsible for controlling the motion of the cobot in one axis ($y-axis$) via a keyboard, while the dRL agent controls the motion of the cobot in the perpendicular axis ($x-axis$); by combining the motions of the two partners the EE  moves in the $xy$ plane (plane of the table's surface). The task is successfully completed if the team manages to drive the cobot's EE within a circle of $r=0.01m$ radius around the goal position while the speed of the EE is lower than than $0.05m/s$. A \textit{game} is completed either if the goal is successfully reached or if 30 seconds expire.

\textbf{Controlled variable.} In the HRC task, both the human and the dRL agent control the robot's acceleration. Specifically, they can provide three discrete actions: 1) a positive acceleration of $+\alpha_{c}$, 2) a negative acceleration of $-\alpha_{c} $ or 3) no (0) acceleration, where $\alpha_{c} > 0$ is a positive constant. This kind of control makes the task challenging for the human; collaborative learning is necessary to achieve the goals of the task.

\subsection{Reinforcement Learning agent}\label{sec:rl}
The motion of the robot in the x-axis is controlled by a discrete dRL SAC agent \citep{lygerakis2021,christodoulou2019soft}. A four-dimensional state space comprises the position $(x_{ee},y_{ee})$ and the velocity $(\dot{x}_{ee}, \dot{y}_{ee})$ of the cobot's EE:

\begin{equation}
    s=(x_{ee},y_{ee},\dot{x}_{ee},\dot{y}_{ee})
\end{equation}

The actions of the human are taken into account in the state space through the kinematic variables of the EE's movement in the $y-axis$. The dRL agent's action space is 1-dimensional and discrete $\alpha=\{-1, 0, 1 \}$. The actions of the agent correspond to the negative, zero, or positive acceleration of the robot's EE in the $x-axis$. Finally, a sparse reward function \citep{shafti2020real} is used, where the agent is penalised with $-1$ at each timestep and rewarded with $10$ if the goal is reached.

Two different  kinds of dRL agents are used in this work. The one agent does not use any transferred knowledge and the selected action is derived from the current SAC policy. The other agent, on the other hand, transfers the knowledge of an expert team. Specifically, this agent selects actions either from the current SAC policy or from the policy of a pre-trained `expert' agent. A probabilistic policy reuse (PPR) method is used for the transfer learning. The actions from the `expert' agent are selected according to a probability $\psi_{ppr}$ as follows:  

\begin{equation}\label{eq:tl}
\alpha = \begin{cases} 
      \arg \max_{\alpha} \pi_{SAC}(\alpha | s),& \text{if } \psi \leq \psi_{ppr}\\
      \arg \max_{\alpha} \pi_{expert}(\alpha | s), & \text{otherwise}  \\
   \end{cases}
\end{equation}
 
where $\psi$ is a random number in the interval [0, 1], $\psi_{ppr}$ is the PPR threshold $\pi_{SAC}$ the current SAC policy and $\pi_{expert}$ the policy of the expert agent. Note that before any off-line dRL training a naive agent is used and its actions are randomly selected. 

The `expert' agent used for applying the PPR was trained by one of the authors. Naturally, the `expert' human collaborator had both experience in the collaborative task and was aware of the control variables. The `expert' is estimated to have had approximately 10 hours of experience with the task.

\subsection{Human-Robot Collaborative study}\label{sec:experiments}
\subsubsection{Participants}
Seventeen participants (age: $mean=29.2$ ($SD=4.8$), 9 female) were recruited for the HRC study. Almost half of them ($43.8\%$) had experience with the AI (in graduate and post-graduate level), while only a quarter of them had experience with robotics. The study protocol was approved by the Research Ethics Committee of the National Centre of Scientific Research `Demokritos'. All participants provided written informed consent. 

The participants were randomly assigned in one of two groups: 9 participants collaborated with the `no transfer learning' agent ($No\_TL$ group) and 8 participants with the agent that used PPR ($PPR$ group). One participant was excluded from the $No\_TL$ group as the behaviour of this human-robot team was considered to be an outlier (see section \ref{Res:TrainNTestGames}). The participants were unaware of the group they were assigned to. Moreover, they were never explicitly told that they would be collaborating with an AI agent. 

The participants were seated in the same table where the cobot was set-up (figure~\ref{fig:setup}) on a height-adjustable stool. They were instructed that they would use a keyboard and \textit{collaborate with the UR3 cobot} in order to move its EE to a goal position. It was explained to them that they were responsible for controlling the motion of the EE towards or away from their bodies, while the robot would control the other direction. Specifically, the following instructions were provided regarding the use of the keyboard: 

\begin{itemize}
    \item By pressing the `i' key you can drive the EE away from you.
    \item By pressing the `,' key you can bring the EE towards you.
    \item By pressing the `k' key you command the EE to keep moving in the exact same way as in this very moment. 
\end{itemize}

Note that pressing the `i' and `,' keys multiple times would not apply extra positive or negative acceleration respectively. Only the change from the one state to the other mattered. Pressing the `k' key was akin to removing any acceleration and maintaining current EE velocity.

At the beginning of each game, once the EE reached one of the four randomly assigned initial positions, a sequence of beeps (three short and one long) signaled the start of the game. A different  auditory feedback was provided depending on the outcome of the game. Furthermore, the outcome (win or lose) as well as the score of the team and the number of the game was provided to the participants in a computer monitor. Finally, the participants were told that the robot movements were constrained within the rectangle area. The experimenter also showed to them the safety button on the teach-pendant of the UR3 and explained that he could press it in case of an emergency to stop the cobot's motion. 

To familiarize with the set-up, the participants were asked to play 7 games during which the participant had to control the EE's movement alone along the $y-axis$ (the axis controlled by the human). At the beginning of each game, the EE was automatically placed in a single initial position (marked as SIP in figure~\ref{fig:setup}). Once the trial started, the participant had 10 seconds to bring the red laser dot into the goal position with a relatively slow speed. The goal position and the positional tolerance were the same as in the HRC game but the velocity tolerance was set to 0.02m/s. The familiarization games also allowed to observe the baseline performance of the participants in controlling the cobot's EE without the involvement of a dRL agent. All of the participants managed to successfully complete the familiarization task at least two times while the 87.5\% of them reached the goal in time at least 4 times. No participants were excluded based on this procedure. 

\subsubsection{Study design}
The study design is shown in figure~\ref{fig:pipeline}. The human-robot teams had to complete 150 games. The games were grouped in 15 \textit{batches}. During the \textit{baseline testing} batch the participants of both groups had to collaborate with a random agent. The rest 14 batches were organised in 7 experimental blocks that comprised:

\begin{figure}[h]
    \begin{center}
    \includegraphics[width=0.8\textwidth]{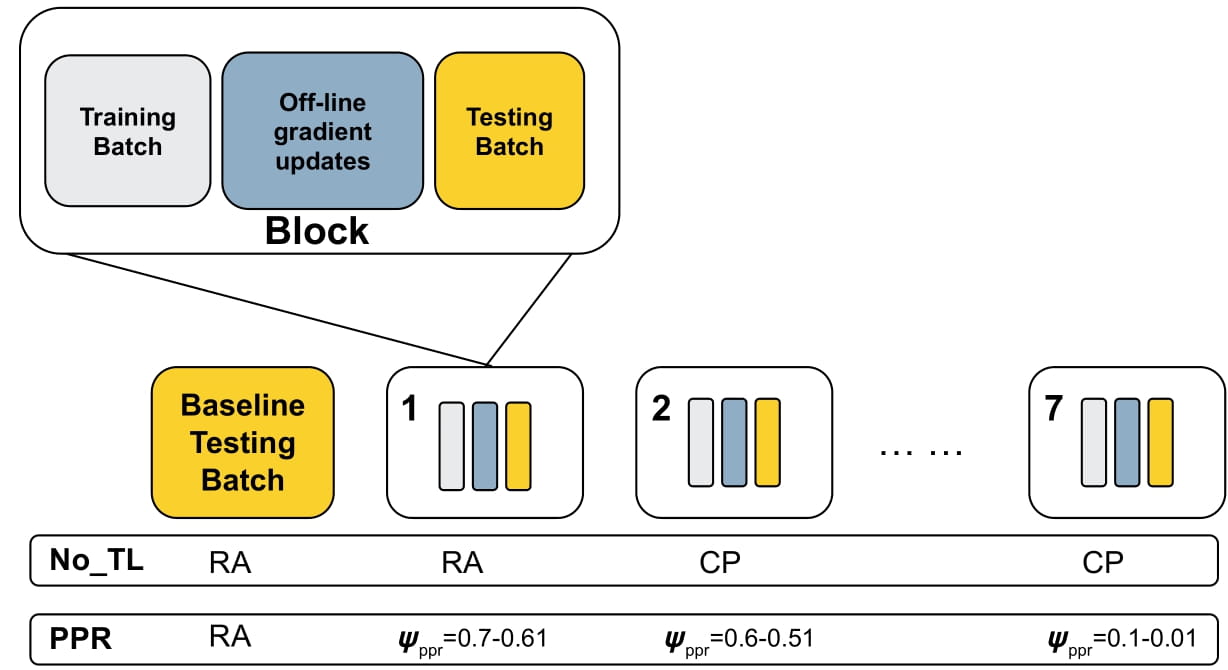}
    \end{center}
    \caption{Study design. Both groups completed a baseline testing batch during which the the RL agent selected random actions (RA). This was followed by 7 experimental blocks that comprised: a training batch (grey boxes), 14K off-line SAC agent gradient updates (blue boxes), and a testing batch (yellow boxes). $No\_TL$: In the first training batch, the agent chose random actions and for the rest of the blocks it selected an action based on the current policy (CP). $PPR$: Additionally to the above, the PPR action selection procedure (Eq.\ref{eq:tl}) was applied with the $\psi_{ppr}$ initially set at $0.7$ and discounted by $0.1$ in each game.}
    \label{fig:pipeline}
\end{figure}

\begin{itemize}
    \item One \textit{training} batch; During the 10 training games the agent's actions depended on the group ($No\_TL$ or $PPR$). In the $No\_TL$ group, the agent selected random actions in the first block and it then sampled from its current policy for the rest 6 blocks. In addition to the above, in the $PPR$ group, the agent chose actions from the `expert' policy according to the probability $\psi_{ppr}$ (see Eq.~\ref{eq:tl}). The value of $\psi_{ppr}$ was set to 0.7 for the first game and decayed by $\psi_\epsilon = 0.01$ in each following game. All of the training games were stored for the off-line SAC training in a buffer of size 1000000.
    \item Off-line SAC agent training; This included 14K gradient updates and lasted approximately 3 minutes. The participants were told that a short break is provided every 20 games. 
    \item One \textit{testing} batch; During the 10 testing games the agent was sampling actions from its current policy for both the $No\_TL$ and the $PPR$ groups. The performance in these batches was used for the evaluation of the two groups. 
\end{itemize}
 Therefore, there were 8 testing and 7 training batches in total. 

 \subsubsection{Collaboration measures}
Both objective and subjective measures were used to evaluate the effect of the TL in the learning and overall performance of the human-robot team in the collaborative task.

\textbf{Objectives measures}. The time needed for completing the study was used as an indicator of the \textit{total training time}. Moreover, the \textit{travelled distance} of the EE during a game was used to evaluate the efficiency of a human-robot team. The travelled distance was multiplied by the percentage of the total available time spent in a game. This \textit{normalized travelled distance} was used to account for the games that the EE was driven to a border point and the team never managed to bring it back into the game, providing an erroneously short distance.

\textbf{Subjective measures}. Several subjective measures were also used to assess the experience of the participants during the collaboration. Table \ref{tab:subj-measures} presents the subjective fluency metric scales that were used. The \textit{fluency}, \textit{contribution}, \textit{trust} and \textit{improvement} metrics were used as described in \cite{hoffman2019evaluating}. The positive teammate traits and the alliance were based on \cite{hoffman2019evaluating} but with some modifications; the cooperative trait was added into the positive team traits and only a subset of the alliance questions were used based on the characteristics of the present study. The \textit{safety} measure for physical systems was used as suggested in \cite{dragan2015effects}. Finally, the participants were also asked to evaluate their own improvement over the game (\textit{self-improvement} - `my performance improved over time').  

\begin{table*}[h]
    \centering
    \caption{Subjective measures} 
    \begin{tabular}{l}
    \hline
    \rowcolor{SpiritBlue}
    \textbf{Human-Robot fluency} $\alpha=0.901$\\
     The human-robot team worked fluently together. \\
     The human-robot team’s fluency improved over time.\\
     The robot contributed to the fluency of the team interaction.\\
    \hline
    \rowcolor{SpiritBlue}
    \textbf{Robot contribution} $\alpha=0.282$\\
    I had to carry the weight to make the human-robot team better. (R)\\
    The robot contributed equally to the team performance.\\
    I was the most important team member on the team. (R)\\
    The robot was the most important team member on the team.\\
    \hline
    \rowcolor{SpiritBlue}
    \textbf{Trust} $\alpha=0.834$\\
    I trusted the robot to do the right thing at the right time.\\
    The robot was trustworthy.\\
    \hline
    \rowcolor{SpiritBlue}
    \textbf{Positive teammate traits} $\alpha=0.856$\\
    The robot was intelligent.\\
    The robot was trustworthy.\\
    The robot was committed to the task.\\
    The robot was cooperative.\\
    \hline
    \rowcolor{SpiritBlue}
    \textbf{Improvement} $\alpha=0.91$\\
    The HR team improved over time.\\
    The human-robot team’s fluency improved over time.\\
    The robot's performance improved over time.\\
    \hline
    \rowcolor{SpiritBlue}
    \textbf{Safety} $\alpha=0.582$\\
    I feel uncomfortable with the robot.(R)\\
    I feel safe working next to the robot. \\
    I am confident the robot will not hit me as it is moving.\\
    \hline
    \rowcolor{SpiritBlue}
    \textbf{Working Alliance for H-R teams [selection]}  $\alpha=0.732$\\
    I am confident in the robot’s ability to help me.\\
    The robot and I trust each other.\\
    The robot perceives accurately my goals.\\
    The robot does not understand what I am trying to accomplish.\\
    I find what I am doing with the robot confusing.\\

    \hline
    \end{tabular}
    \label{tab:subj-measures}
\end{table*}
In addition to the collaboration measures mentioned above, the participants were asked to give a \textit{judgment of control} during the game using a question based on \cite{dewey2014phenomenology}: 
``\textit{How would you rank your ability to control the motion of the robot in the last 10 games from 1 (no control) to 9 (perfect control)? Consider that you are evaluating how efficient the use of a keyboard was to control the robot and how much your actions contributed to the outcome of each game.}'' Only this questions was posed to the participants three times after the $1_{th}$, $4_{th}$ and $7th$ training batches. The rest of the subjective measures were only evaluated once at the end of the study. Note that all questions were posed in participants' native language (Greek).

\subsection{Robotic set-up}\label{tech:robot_control}

\subsubsection{Robot control}
Both the human and the dRL agent controlled the motion of the EE in their respective axis by providing commanded accelerations. The commanded accelerations were then numerically integrated to commanded velocities. Briefly, the control design consists of a feedforward term on the acceleration as follows:

\begin{align}\label{eq:control}
    \Ddot{\textbf{x}}_{com} &= u \\
     \dot{\textbf{x}}_{com} &= \dot{\textbf{x}}_{com}+ T_c * \Ddot{\textbf{x}}_{com}
\end{align}

where $u$ is the desired acceleration imposed by the human or the RL agent, $\Ddot{\textbf{x}}_{com}$ is the commanded acceleration, $\dot{\textbf{x}}_{com}$ is the commanded velocity and $T_c$ is the control cycle. In our case $T_c = 0.008s$ because the robot controllers operate at 125Hz.

The motion of the robot is regulated by commanded EE velocities. The EE velocities are then mapped to joint velocities using Inverse Kinematics. The joint velocities are then passed to the robot controllers. 

\subsubsection{ROS Integration}

The entire system was integrated into ROS and run on Melodic and Ubuntu 18.04. The main components of the ROS pipeline are the following: 
\begin{itemize}
    \item \textit{Human command input}: The human controls the motion of the EE in the $y-axis$ through a keyboard. A node listens to the keyboard input and publishes ROS messages which correspond to the human desired acceleration. Specifically, `i' button applies an acceleration of $+0.4 \: m/s^2$, the `,' button an acceleration of $-0.4 \: m/s^2$ and the `k' button applies zero acceleration. These values were chosen experimentally.
    \item \textit{RL command and game loop}: A node provides the action of the agent and the loop of the game. The control of the motion in the $x-axis$ is the same as in the $y-axis$. The agent takes a new action, sampled from the policy provided by the SAC algorithm every 200ms. A discrete action SAC \footnote{\url{https://github.com/Roboskel-Manipulation/maze3d_collaborative}} was used in the RL loop.
    \item \textit{Robot motion generation}: A node listens to the human and the agent commanded accelerations and implements the feedback control law described in Section~\ref{tech:robot_control}. Furthermore, this node also restricts the motion of the robot inside the square. The EE position in both $x-$ and $y-axis$ is checked and in case it is in maximum or minimum position, the EE commanded velocities become zero. The EE starts moving once again depending on the sign of the commanded acceleration. This procedure gives a feeling to the human participant as if the EE hits a virtual wall. Finally, the output of the robot motion generation node (EE velocities) is mapped to joint velocities through Inverse Kinematics that are then executed by the robot controller.
\end{itemize}

%%%%%%%%RESULTS%%%%%%%%%%%%

\section{Results}
\subsection{The HRC game}

\begin{figure}[h]
    \begin{center}
    \includegraphics[width=0.9\textwidth]{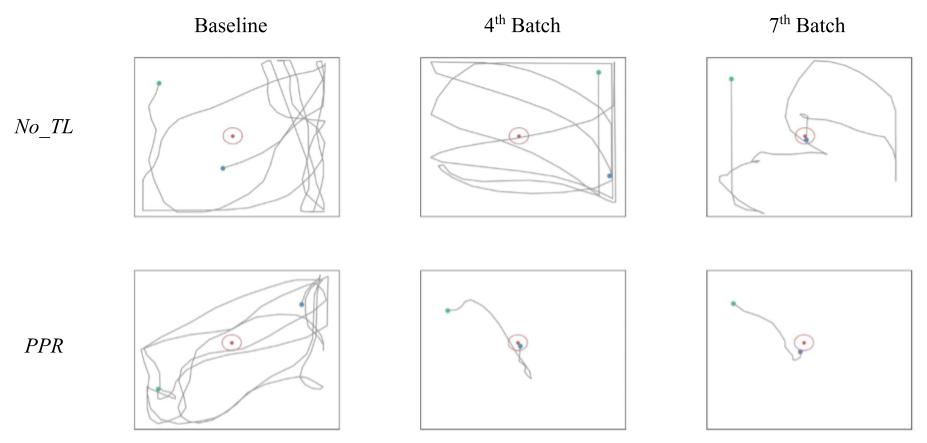}
    \end{center}
    \caption{Single game behaviour of a team in the $No\_TL$ group ($1^{st}$ row) and a team in the $PPR$ group ($2^{nd}$ row). The figures show the EE's travelled paths in the Baseline, $4^{th}$ and $7^{th}$ testing batches, from the starting position (green dot) to the final position (blue dot). The goal area is shown with the red circle.}
    \label{fig:Single-Games-behaviour}
\end{figure}

Figure~\ref{fig:Single-Games-behaviour} shows six paths travelled by the EE as a result of the collaboration. These paths demonstrate a typical behaviour of two teams in the $No\_TL$ and $PPR$ groups in single games across the study. In the \textit{Baseline} batch games, both teams failed to collaborate successfully and they did not reach the goal; the final point of the path is outside the target area. However, the improvement in the task for the $PPR$ team is evident already in a game of the $4^{th}$ batch. In the games of the $7^{th}$ batch both teams successfully complete the task. However, the $PPR$ team also drives the EE to the target through a shortest path.

\begin{figure}[h]
    \begin{center}
    \includegraphics[scale=0.20]{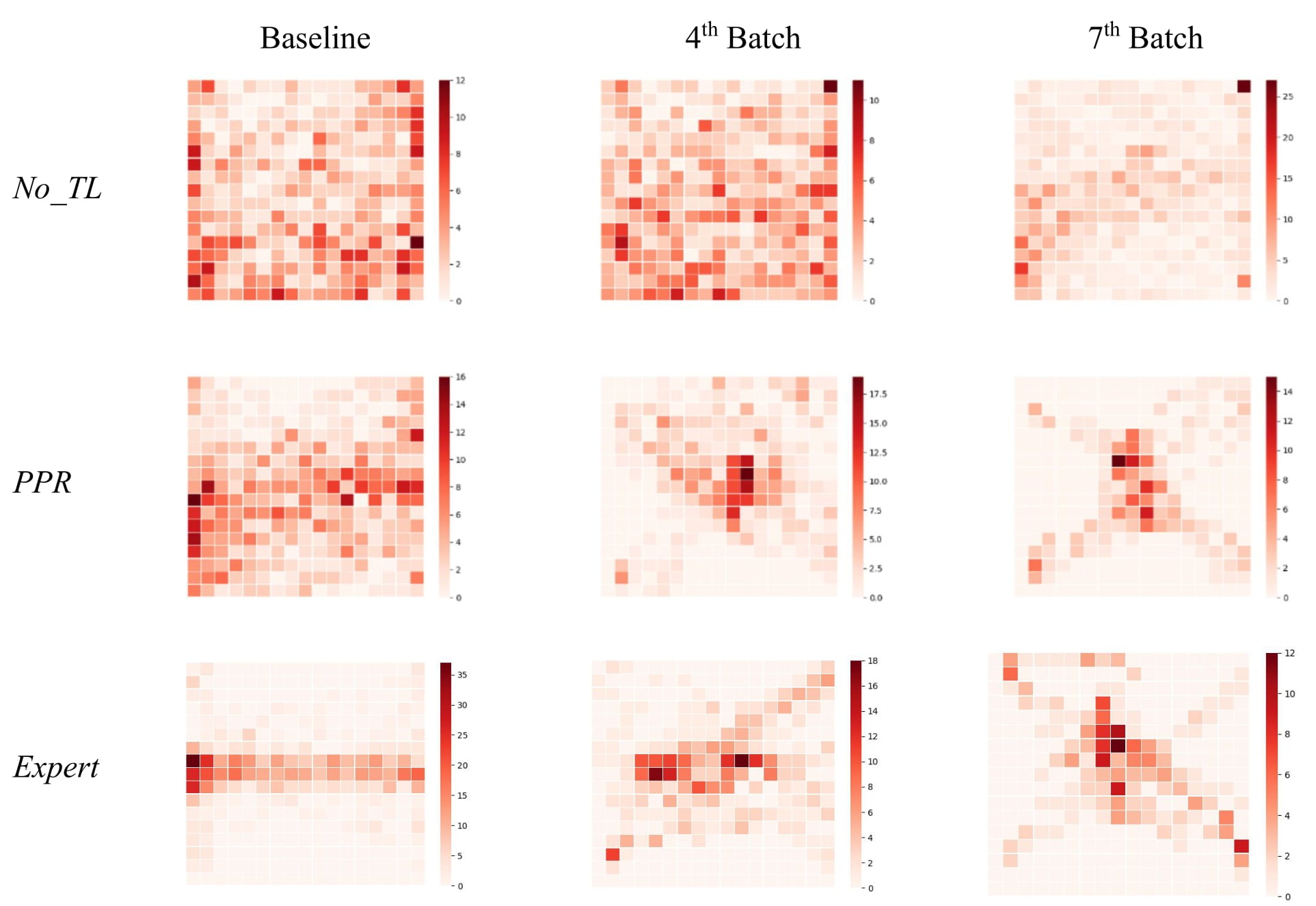}
    \end{center}
    \caption{Testing batch behaviour of a team in the $No\_TL$ group ($1^{st}$ row), a team in the $PPR$ group ($2^{nd}$ row) and the team with the expert human ($3^{rd}$ row). The heatmaps show the laser dot's position in the Baseline, $4^{th}$ and $8^{th}$ testing batches. The numbers indicate the frequency with which the dot occupied each cell ($1cm\times1cm$) - that is the number of $x,y$ pairs counted within the cell in all ten games of a batch.}
    \label{fig:Batches-behaviour}
\end{figure}

Similar observations are derived by the heatmaps in figure~\ref{fig:Batches-behaviour} that show the behaviour of two teams in the $No\_TL$ and $PPR$ groups across the different batches of the study. The heatmaps show the occupancy frequency of each cell of the workspace for all ten games of each batch. During the baseline games the majority of the cells are occupied for both teams. However, in the $PPR$ condition this changes over the time and during the final ten games the team drives the EE almost through a line connecting the start points and the goal. This is demonstrated by the `X'-shaped occupancy grid. In the case of the $No\_TL$ team, an incline towards the left side of the space is observed signifying that the RL agent learnt to lead the EE towards the left of the workspace. Note that the highly occupied cells close to the starting positions (as in $4^{}th$ and $7^{}th$ $No\_TL$ batches) result from games were the EE had stuck in a certain position, while the highly occupied cells close to the centre (as in $4^{}th$ and $7^{}th$ $PPR$ batches) result from successful target hits. All in all, the figure indicates that using PPR enables the team to win while travelling shorter distances compared to the $No\_TL$ case. The last row of figure~\ref{fig:Batches-behaviour} also demonstrates the behaviour of the team with the expert human. The baseline batch mirrors the experience of the human; the expert human managed to drive and maintain the EE in a horizontal zone around the target. The variability along the $x-axis$ demonstrates the behaviour of the naive agent. By the end of the training the `X'-shaped occupancy grid was achieved.

\subsection{Training and testing games}\label{Res:TrainNTestGames}
\begin{figure}
     \centering
     \begin{subfigure}[b]{0.45\textwidth}
         \centering
         \includegraphics[width=0.9\linewidth]{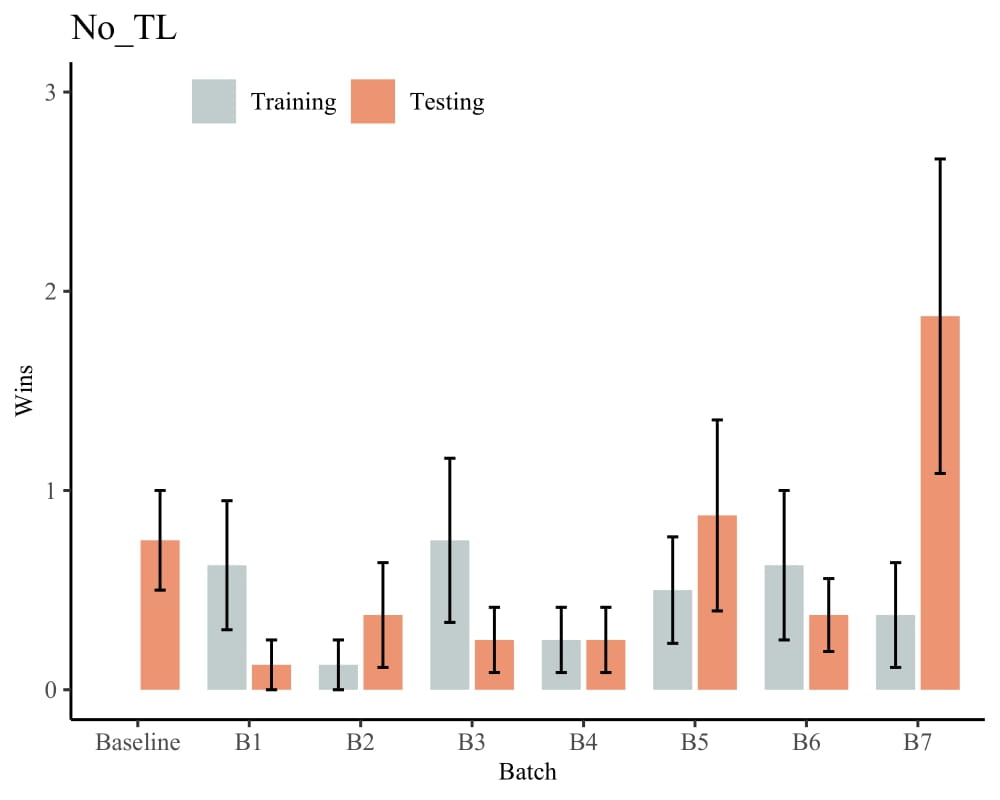}
         \caption{}
         \label{fig:y equals x}
     \end{subfigure}
     \hfill
     \begin{subfigure}[b]{0.45\textwidth}
         \centering
         \includegraphics[width=0.9\linewidth]{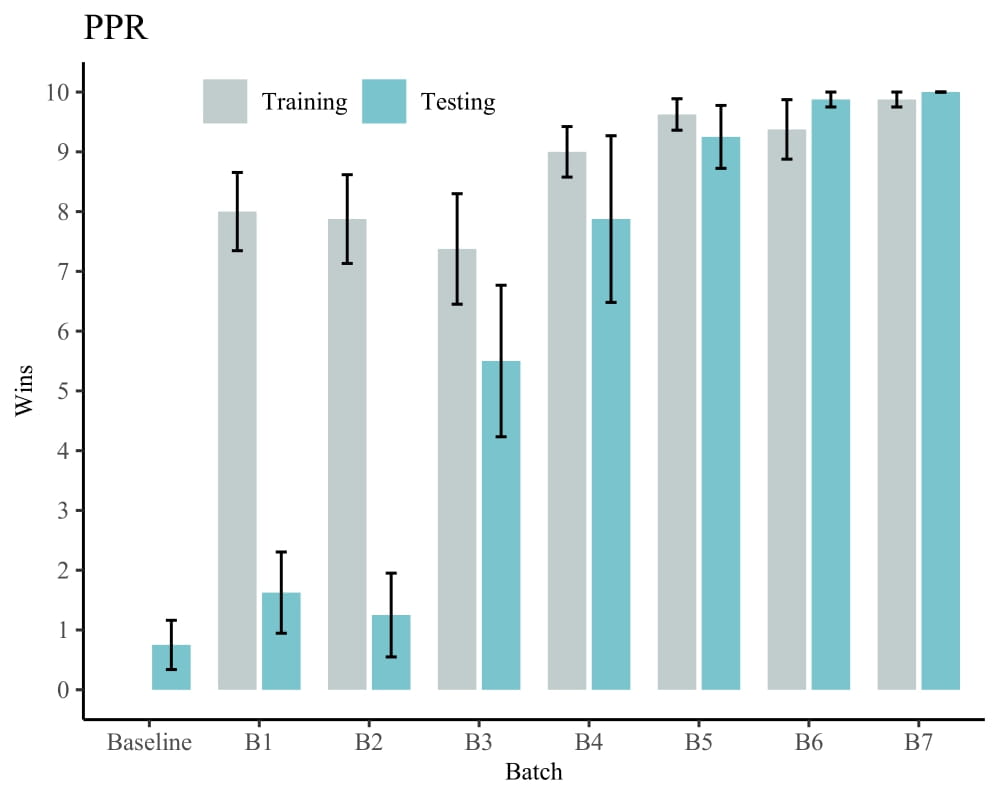}
         \caption{$y=3sinx$}
         \label{fig:three sin x}
     \end{subfigure}
     \hfill
        \caption{Number of wins (successfully completed games) in both the training and the testing batches of the  $No\_TL$ (\textbf{left}) and the $PPR$ groups (\textbf{right}). The errorbars represent the standard error of the mean.}
        \label{fig:wins}
    
\end{figure}

Figure \ref{fig:wins} shows the average number of wins in both groups across both the training and the testing batches. Although both groups present a similar behaviour of low wins in the baseline testing batch (note the difference in the range of units in the two subfigures), the effect of the TL in the $PPR$ group is evident already from the first training batch. While the $PPR$ group succeeded to reach the goal 8 times on average, the behaviour of the $No\_TL$ group was no different from the baseline batch as a result of the random agent. Overall, the result of the TL can be observed in the difference in the successful games across the training batches (grey bar comparison in figure \ref{fig:wins}) between the two groups.  Naturally, the training batches affect the off-line SAC gradient updates and consequently the resulting policy. In the case of the $PPR$ team, the performance is considerably boosted due to the expert knowledge transferred, whereas for the $No\_TL$ teams it merely depends on chance. Actually, one $No\_TL$ human-robot team happened to succeed in three games during the first batch, and by the $4^{th}$ batch it had accumulated 25 wins in the 40 games. At the same time, the next best score in the group was 5 wins. The behaviour of this team was an outlier and was excluded from any analysis.

\begin{figure}
    \begin{center}
    \includegraphics[scale=0.15]{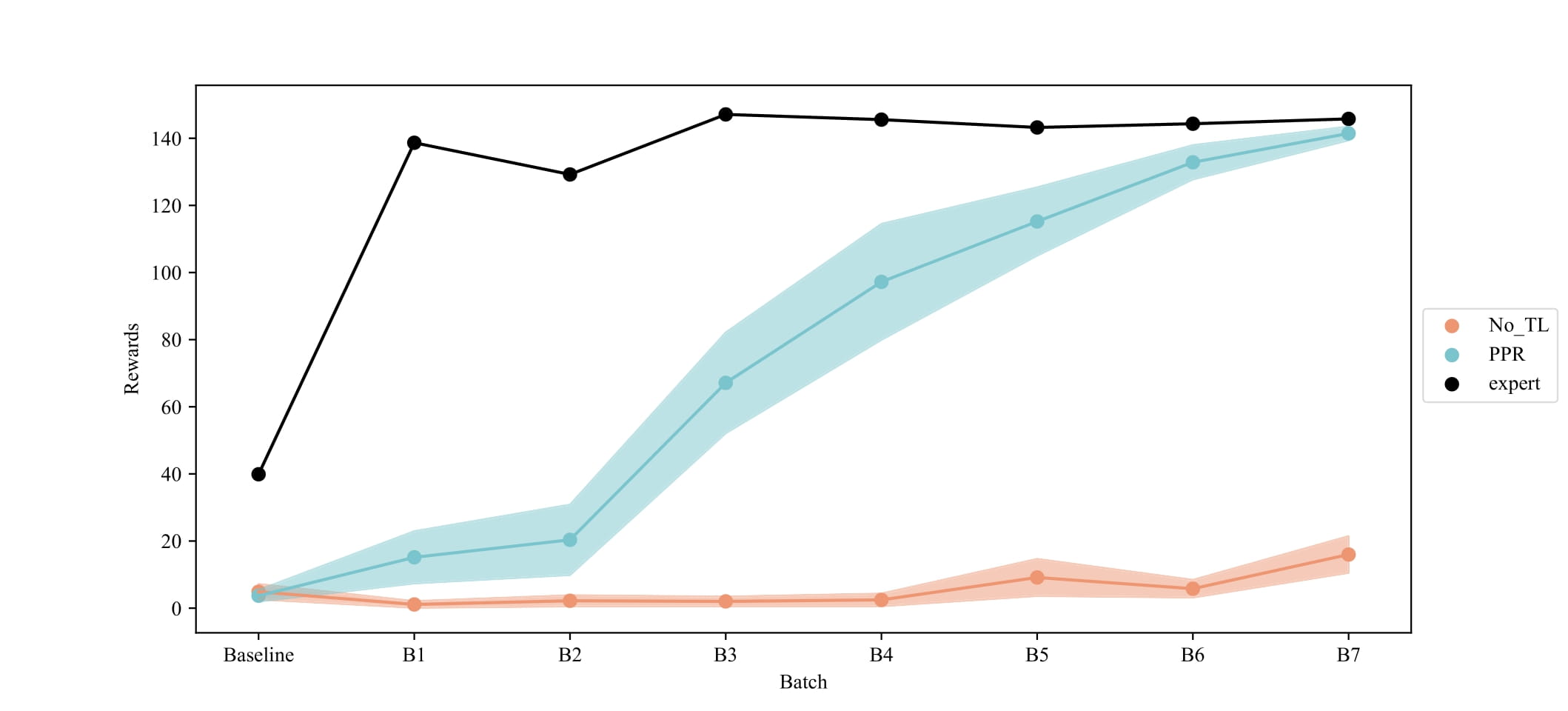}
    \end{center}
    \caption{Average group \textit{rewards} across the testing batches for the $No\_TL$ (red) and $PPR$ (blue)groups. The expert's learning curve is also shown (black points). The shaded areas represent the standard error of the mean.}
    \label{fig:rewards}
\end{figure}

The effect of the TL in the learnt policy is shown both in the number of average wins per testing batch (figure \ref{fig:wins}) but also in the average received reward in these batches (figure \ref{fig:rewards}). It is evident that TL assisted the human-robot teams in achieving a behaviour similar to the expert by the end of the study. The human-robot teams managed to win in almost all the games. Moreover, based on the obtained rewards, it took on average approximately 4 seconds to complete a game. In the case of the $No\_TL$ group, the teams achieved on average less than 2 wins and in these cases they consumed almost all the given time (a reward of $+$10 means that the teams reached the goal just right at the end of the game). Note that the team with the expert human, achieved on average considerably higher rewards even in the baseline batch as a result of the expert human's experience. The team managed to win 5 times. In the batches that followed, although the expert human collaborated with a $No\_TL$ agent the team successfully completed all ten testing games in an efficient way.

\subsection{Objective collaboration measures}
\begin{figure}[ht]
    \begin{center}
    \includegraphics[scale=0.15]{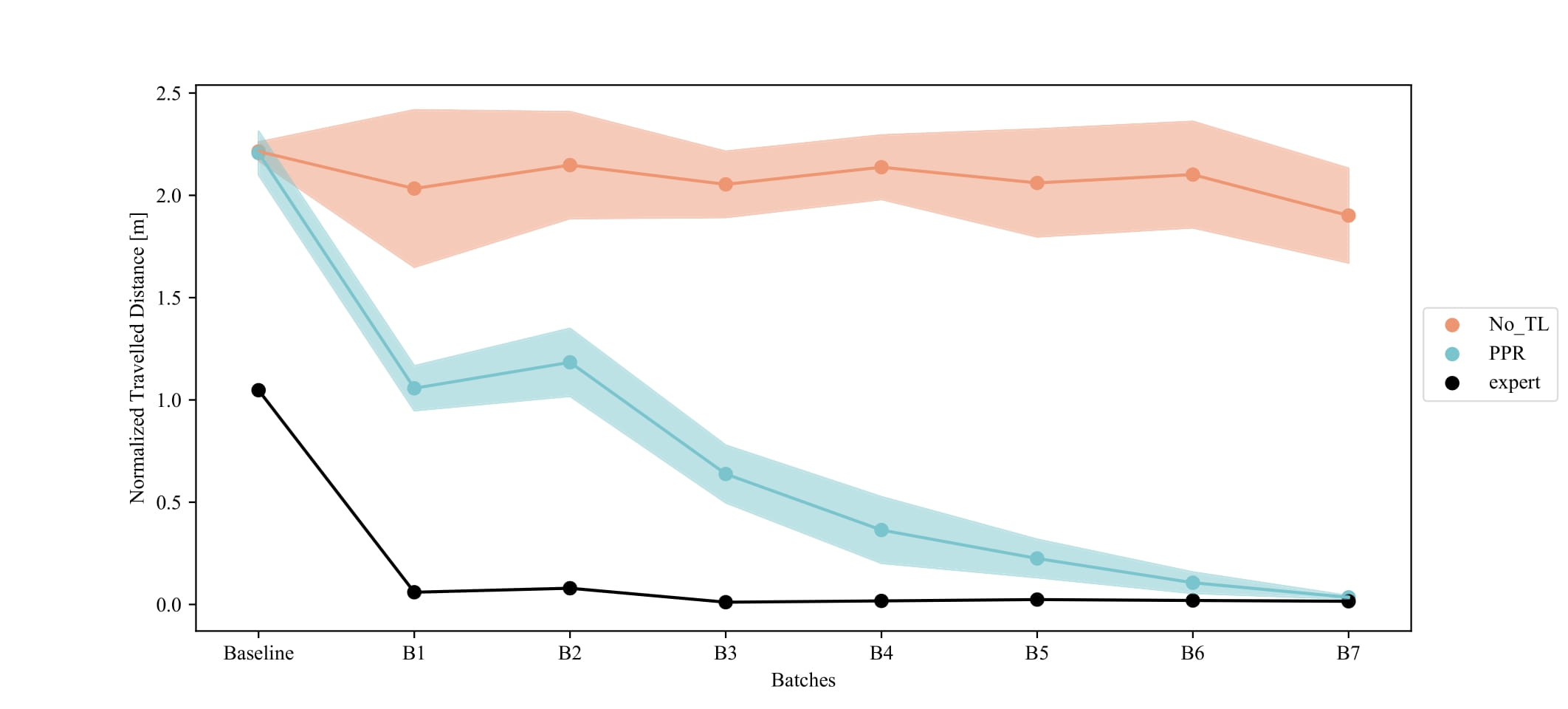}
    \end{center}
    \caption{Average \textit{normalized travelled distance} across the testing batches for the $No\_TL$ (red) and the $PPR$ (blue) groups. The expert's learning curve is also shown (black points). The shaded areas represent the standard error of the mean.}
    \label{fig:travelled_distance}    
\end{figure}

 The observations described above were evaluated both through the \textit{total training time} (total duration of the testing and training batches) and the \textit{normalized travelled distance}. The total training time was significantly lower for the $PPR$ group compared to the $No\_TL$ group ($t(7.6) = 11.7$, $p < 0.001$). Specifically, it took on average 73.1 minutes ($SD=1.94$) for the $No\_TL$ teams to complete the study while the $PPR$ needed on average 33.7 minutes ($SD=9.32$). The training of the team with the $No\_TL$ agent and the expert human lasted 15.81 minutes. Figure \ref{fig:travelled_distance} shows the normalized travelled distance of the two groups and the expert. 

The observations are similar to those of the total wins and the received rewards. The $PPR$ teams not only manage to successfully reach the goal but they did so by driving the EE though a path close to the shortest distance ($d=12cm$) between the starting and goal positions.  To confirm the differences \emph{between} the two groups  and \emph{within} each batch a \emph{robust} mixed ANOVA (`bwtrim' function in WRS2 R package~\cite{mair2019robust}) was used as Shapiro-Wilk normality test for two batches was significant. The analysis showed a significant main effect of \emph{group} ($Q(1,7.25)=42.51$,  $p<0.001$) and of \emph{batch} ($Q(7, 6.59)=6.02$, $p=0.0175$). Moreover, there was also a significant interaction of the \emph{group$\times$batch} ($Q(7,6.59)=5.32$, $p=0.024$). 

\subsection{Subjective collaboration measures}
In addition to the objective measures, the participants' experience during the collaboration was also studied. Table~\ref{tab:subj-measures} lists the subjective measures used in the present study. The Cronbach's $\alpha$ is also reported there. The fluency, trust, positive teammate traits, and the improvement scales had good or excellent internal consistency, and the alliance scale an acceptable one. However, the safety had a poor internal consistency. Upon inspection of the data, it was observed that all participants from both groups had `completely agreed' (7 on the Likert rating) with statement about being confident that the robot will not hit them (this is discussed in the next section). After dropping this item the internal consistency of the scale improved to acceptable ($\alpha=0.776$). So, in the analysis reported below, only the first two items of the scale are included. Finally, the  contribution measure had an unacceptable internal consistency and was examined separately. 

\begin{figure}[h]
    \begin{center}
    \includegraphics[width=0.6\textwidth]{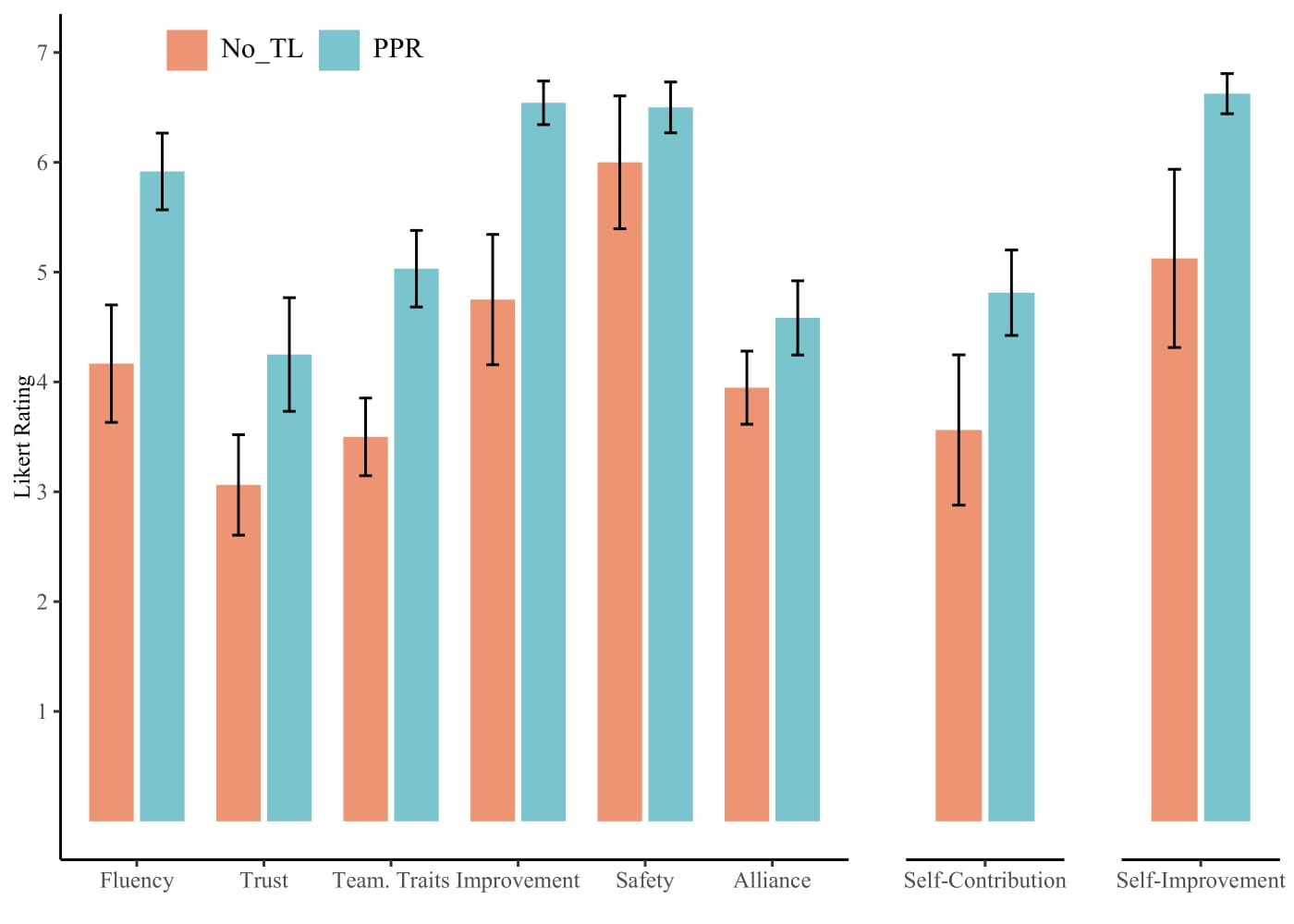}
     \end{center}
    \caption{The subject collaboration measures for the $No\_TL$ (orange) and the $PPR$ (blue) groups.}
    \label{fig:subj-measures-errorbars}
\end{figure}

The results of the fluency, trust, positive teammate traits, improvement and alliance scales are shown in figure \ref{fig:subj-measures-errorbars}. Although there is a tendency towards higher ratings in all the scales for the $PPR$ group only the fluency ($p=0.0178$), positive teammate traits ($p=0.008$) and improvement ($p=0.0145$) were significant. Regarding trust,  participants in both groups provided relatively low ratings (compared to the other scales). Safety was rated quite high by both groups. 

After exploring further the contribution scale, it was found that the two questions regarding the robot contribution yield a negative internal consistency. This can potentially be explained by the fact that some participants seemed confused about whether these questions regarded the robot behaviour during or towards the end of the study. Moreover, considering the nature of the present study and the inefficient performance of the $No\_TL$ group some participants might have rated the statement regarding the importance of the robot in terms of `the robot having the most important contribution towards the \textit{failure} of the game'. On the other hand, the items that concerned the participants' \textit{self-contribution} had a good internal consistency ($\alpha=0.818$). The self-contribution scale is also shown in figure \ref{fig:subj-measures-errorbars} without reversed the ratings. The difference between the two groups in their rating of self-contribution was significant ($p=0.011$). Finally, participants in both groups considered that their performance improved over the time ($p=0.27$).

\begin{figure}[h]
    \begin{center}
    \includegraphics[scale=0.20]{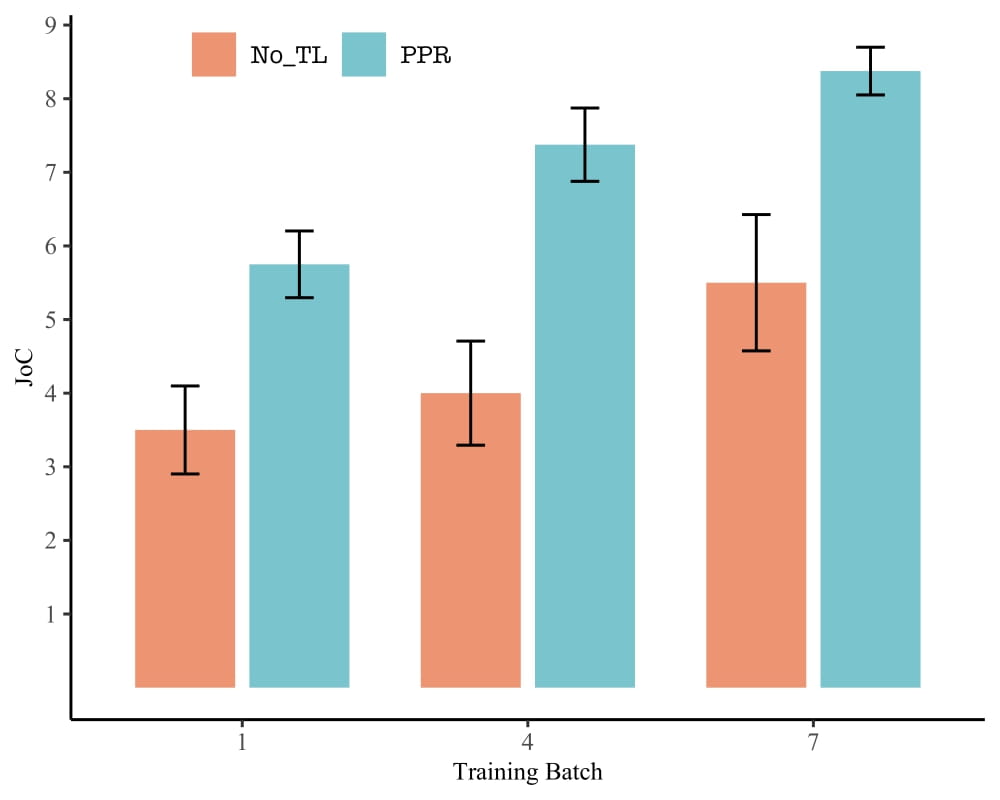}
    \end{center}
    \caption{The judgment of control (JoC) ratings for the $No\_TL$ (orange) and the $PPR$ (blue) groups after the $1^{st}$, $4^{th}$ and $7^{th}$ training batches. }\label{fig:subj-measures-JOC}
\end{figure}

Finally, figure \ref{fig:subj-measures-JOC} shows the judgement of control (JoC) of the participants in both groups and across the different batches. A two-way repeated ordinal analysis of variance (clmm function in R) showed that there was a significant effect of group ($\chi^2=9.29$, $p=0.002$) and batches ($\chi^2=27.5$, $p<0.001$), but not a significant interaction ($\chi^2=1.87$, $p=0.39$).

The first time the participants were asked about the JoC was just after the first training batch. As the $PPR$ groups had already been exposed to the PPR agent, there is a trend towards higher ratings. However, a post-hoc analysis showed that the difference between the two groups is not significant ($p=0.06$).  Note that, it was decided not to ask this question just after the baseline batch to avoid interrupting the session of the first two batches. The difference in the JoC ratings at the end of the study within each group significantly increased between the first and the last rating of JoC ($No\_TL$: $p=0.01$, $PPR$: $p<0.001$). Moreover, both after the $4^{th}$ training batch and at the end of the study the JoC difference between the two groups was significant (after the $4^{th}$ batch:$ p=0.0045$, after the $7^{th}$ batch:$p=0.0047$).

Finally, a series of ordinal-to-ordinal (polychoric) correlations were calculated to explore the relationship between the JoC and the objective and subjective measures. A positive correlation (0.885) indicates a strong positive relationship between the JoC at the end of the training (JoC$_{7}$) and the scale of improvement. Similarly, a high polychoric correlation exists  between the JoC$_{7}$ and the rating of the self-improvement (0.892). A lower degree of correlation seems to exist between the JoC$_{7}$ and the self-contribution ratings (0.61). Both the rating of the self-improvement and the JoC$_{7}$ appear to have a weak positive correlation to the objective measures of total wins in the last testing batch.

%%%%%%%%DISCUSSION%%%%%%%%%%%%

\section{Discussion}
The goal of the present study was to enhance the performance of human-robot teams during collaborative learning by transferring knowledge from an expert team. The results confirmed the initial hypothesis that TL would significantly affect the learning procedure. Specifically, the $PPR$ teams managed to successfully complete the task in most of the games by the end of the training. In addition, they did so in an efficient way that is by travelling close to the shortest path. As expected this massively affected the total training time; the $PPR$ group completed the overall training procedure (on average) in less than half the time compared to the $No\_TL$ group (73.1 and 33.7 minutes respectively). Finally, the $PPR$ group approached the expert behaviour by the end of the training as shown by both the average rewards at the last testing batch and the normalized travelled distance. 

Although \cite{shafti2020real} do not report the training times, by considering the average success rates provided and an average time of 20 seconds for each successful trial, we estimated that it should have taken approximately 36 minutes for each participant to complete the 8 training batches. This time is comparable to the $No\_TL$ group in our study (note that the 73.1 minutes in our case account for a double number of batches since both training and testing batches are included). Unlike \cite{shafti2020real}, we did not observe the same degree of variability within participants of each group. This can be explained by both the initial and overall experimental conditions. In the case of the $PPR$ group the expert behaviour was dominant during the first batches and positively affected the overall performance of the teams, possibly masking any individual differences. Moreover, in our case, both groups collaborated with a naive agent at the first batch as opposed to a pre-trained agent. This seems to have had a negative impact to the success of the $No\_TL$ group. A pre-trained agent might have heightened the salience of the between participants differences. In any case, a high success rate during the first trials provides a significant advantage to the final outcome. This was also observed in the present study with the participant that was excluded from the $No\_TL$ group (see section~\ref{Res:TrainNTestGames}). Since no initial `assistance' was provided to our groups, the performance of teams in the $No\_TL$ groups was greatly constrained. 

Apart from the experimental conditions as discussed above, a notable difference between the present study and the study of \cite{shafti2020real} is the nature of the task. In our case a discrete action task was used in contrast to the continuous action `tray task'  of \cite{shafti2020real}. Naturally, the latter is a more demanding task in terms of (human) motor control and learning \citep{krakauer2019motor} and thus it is subject to greater within and between participants variability in the behaviour. Actually, this aspect of the `tray task' in combination with the results of \cite{shafti2020real} with respect to the performance of the participants when collaborating with an agent trained with other subjects suggest that TL should be applied in a way that favors human partner's inherit skills and capabilities. This is especially important in the continuous tasks. The task used in the present study is such that action selection is limited to three states and action execution involves just the pressing of keys at the right time. In this sense, TL is unlikely to constrain personal skills. Nevertheless, the trade-off between generalization of AI knowledge and personalization to a human partner should be carefully considered based on the nature of the task.

Another objective of the present study was to evaluate the collaboration in subjective terms related to the experience and perceptions of the participants. The results showed that TL significantly and positively affected the participants' experience of the team \textit{fluency}. Moreover, the participants in the $PPR$ group rated significantly higher the \textit{positive teammate traits} of the robot as well. Nevertheless, the average Likert rating of the positive teammate traits scale was on average around 5, perhaps indicating that participants were hesitant in attributing the traits asked to the robot. This could be related either to the overall expectations regarding the performance of the robot or the extent to which the participants actually experienced the robot as a teammate during the collaboration. The latter might be supported by the results of the working-alliance scale. There both groups (on average) `did not agree or disagree'  overall with the scale. However, further studies are required to distinguish between the two. The participants in the $No\_TL$ group did not seem to agree about the robot having positive teammate traits.

Transfer learning differently affected the \textit{improvement} rating as well. Specifically, there was a significant difference between the two groups. Note that the scale includes items regarding the improvement of the robot's and the team's performance. The $PPR$ group clearly agreed with the improvement statements (average above 6). Somewhat surprisingly, compare to the objective evaluation results, some participants in the $No\_TL$ group also thought so (average above 4). Actually, when looking at the rating of the \textit{self-improvement}, participants in both groups considered that their performance improved over the time and there was no significant difference between the two groups. Again this result is unexpected considering the objective team performance of the $No\_TL$ group. A possible explanation for this is related to the perceived judgment of control and its correlation to the self-improvement; participants felt that they could use more efficiently the keyboard to control the robot over the time. Nevertheless, a social-desirability bias might also exist. 

Three scales were not affected by the TL: the trust, the safety and the working alliance. With respect to trust, both groups were neutral to negative about trusting the robot. Actually, this scale gathered the lower ratings for both groups. This can be explained by the expected and the actual robot performance, as well as other personal biases or attitudes towards the robot. Further investigation would be required to distinguish between the three. On the other hand, both groups highly rated the safety of the system; this scale obtained the higher ratings for both groups. This is attributed to both the verbal instructions provided at the beginning of the study and the overall experience in the set-up. Although the two collaborators were situated in the same workspace, the robot's motion was confined in a specific area and its control gave the feeling of being constrained in this area by a virtual wall. Actually a participant even commented about this: `\textit{it's as if it} (the cobot) \textit{hits a wall when reaching the limits }(of the square)\textit{...perfect!}'. Finally, both groups were on average close to neutral with respect to the working alliance for the team.  Again, this result cannot be easily interpreted by the data collected in this study. 

An important finding of the present study was the lack of internal consistency of the robot's contribution items. As mentioned earlier in the result section, a possible explanation can be the lack of framing  of the questions in terms of low or high performance. For example, in the case of the $No\_TL$ group, participants could either agree that the robot was the most important team member \textit{`in failing to achieve high performance'} or disagree that the robot was the most important team member \textit{`with respect to the successful games'}. Moreover, there seems to be an imbalance between the items that refer to the human and robot contribution. The  human contribution items are similar in terms of the importance of the human role in two different dimensions: overall contribution and contribution towards the learning. The robot contribution items, however, seem to evaluate different levels of contribution: equal or most important. Based on the present results, in the context of similar studies the contribution scale needs to be reconsidered for improving its internal consistency. 
Finally, with respect to  the judgement of control measure, it was observed that participants in both groups felt more in control across the training blocks, albeit in a significantly different degree already in the middle of the study. Actually, although there was an increase in the ratings of the $No\_TL$ group the average rating of JoC towards the end of the study is just above neutral.  

Overall, the results of subjective the measures presented above show that the agent's performance and errors affect both the performance and the reliance \citep{daronnat2020impact} towards the agent. Moreover, it was observed that participants in the $PPR$ group tended to rate their own contribution and improvement quite high, without acknowledging at the same time the contribution of the robot. This observation bears a significant weight in terms of the transparency of the robot's behaviour and highlights the need of providing some sort of feedback for the agent's behaviour \citep{vouros2022explainable}. Explaining the contribution of the agent could increase the trustworthiness of the robot partner and perhaps the overall sense of collaboration.

A limitation of the present study is that no personal traits questionnaire (such as the `Big-5' questionnaire) was administered to the participants. That was a decision based on constraints for the total duration of the the study. However, personal traits could explain some of the findings regarding the subjective measures. On the other hand, in order to observe such correlations a considerably larger sample of participants might be necessary. 

To sum up, the results of the present study show that transfer learning had a significant effect in the performance of the human-robot teams both in terms of the objective and the subjective evaluation metrics. TL halved the time needed for the training of new participants to the task and it also positively affected the sense of fluency of the human partners. Moreover, it was observed that the improvement and self-improvement variables do not correlate with the overall team performance and seem to be affected by the self judgement of control. Also, participants seem to have failed to bond with the cobot as a collaborator. Taken together, these results suggest that explainability  of the cobot's behaviour and possibly other social cues might be necessary to further promote the social profile of the cobot that integrates the TL agent.

\bibliographystyle{unsrtnat}
\bibliography{references}  %%% Uncomment this line and comment out the ``thebibliography'' section below to use the external .bib file (using bibtex) .





% \end{thebibliography}

\end{document}